\pdfoutput=1

\documentclass[11pt]{article}

\usepackage[preprint]{acl}
\usepackage{multirow}
\usepackage{times}
\usepackage{latexsym}
\usepackage{bbding}
\usepackage{pifont}
\usepackage{amssymb}
\usepackage{amsmath}
\usepackage{booktabs}

\usepackage[T1]{fontenc}

\usepackage[utf8]{inputenc}

\usepackage{microtype}

\usepackage{inconsolata}

\usepackage{graphicx}

\title{DyPCL: Dynamic Phoneme-level Contrastive Learning \\for Dysarthric Speech Recognition}

\author{Wonjun Lee$^*$$^1$, Solee Im$^*$$^2$, Heejin Do$^2$,\\ \textbf{Yunsu Kim}$^3$,  \textbf{Jungseul Ok}$^{1,2}$, \textbf{Gary Geunbae Lee}$^{1,2}$\\
        $^1$Department of Computer Science and Engineering, POSTECH, Republic of Korea \\
	 $^2$Graduate School of Artificial Intelligence, POSTECH, Republic of Korea \\
    $^3$aiXplain Inc., Los Gatos, CA, USA \\
    \texttt{\small \{lee1jun, solee0022, heejindo, jungseul.ok, gblee\}@postech.ac.kr}, \texttt{\small yunsu.kim@aixplain.com}
    }

\begin{document}
\maketitle
\begin{abstract}


Dysarthric speech recognition often suffers from performance degradation due to the intrinsic diversity of dysarthric severity and extrinsic disparity from normal speech. To bridge these gaps, we propose a Dynamic Phoneme-level Contrastive Learning (DyPCL) method, which leads to obtaining invariant representations across diverse speakers. We decompose the speech utterance into phoneme segments for phoneme-level contrastive learning, leveraging dynamic connectionist temporal classification alignment. Unlike prior studies focusing on utterance-level embeddings, our granular learning allows discrimination of subtle parts of speech. In addition, we introduce dynamic curriculum learning, which progressively transitions from easy negative samples to difficult-to-distinguish negative samples based on phonetic similarity of the phoneme. Our approach to training by difficulty levels alleviates the inherent variability of speakers, better identifying challenging speeches. Evaluated on the UASpeech dataset, DyPCL outperforms baseline models, achieving an average 22.10\% relative reduction in word error rate (WER) across the overall dysarthria group.

\end{abstract}

\section{Introduction}

\def\thefootnote{*}\footnotetext{Equally contributed}\def\thefootnote{\arabic{footnote}}

Accurate recognition of dysarthric speech, which is slurred and difficult to understand, is critical for assisting effective communication for individuals with speech impairments \cite{young2010difficulties}. However, due to the inherent diversity of severity levels and substantial differences compared to normal speech, dysarthric speech recognition (DSR) poses significant challenges. Previous studies mainly focused on data augmentation \cite{PranantaH0S22, jiao2018simulating, wang2023duta} and speaker-adaptive training \cite{yu2018development, hu2019cuhk, linexploring}. However, relying on additional augmentation techniques or feature extraction methods limits the practical applicability. 


Contrastive learning has been explored in DSR to learn invariant representations by using healthy speech as a stable reference point \cite{wu2021sequential, proto}. The model can more effectively capture the underlying linguistic content in dysarthric speech by anchoring the learning process on phonetic embeddings from healthy speakers despite surface-level variations. For instance, \citet{wu2021sequential} applied pyramid pooling to distinguish words within audio, while \citet{proto} used word-level contrastive learning with entire audio segments. However, word-level embeddings fail to achieve fine-grained recognition, which is crucial for dysarthric speakers with distinct pronunciation challenges.





In this paper, we propose a dynamic phoneme-level contrastive learning (DyPCL) framework, integrating phoneme-level speech embedding with contrastive learning. 
DyPCL incorporates two-way dynamic approaches: dynamic connectionist temporal classification (CTC) alignment and dynamic curriculum learning.
First, we introduce a dynamic CTC alignment method that accurately aligns speech embeddings with phoneme labels for phoneme-level contrastive learning. Unlike previous approaches that rely on external alignment modules for phoneme-level contrastive learning \cite{scala}, dynamic CTC alignment simultaneously learns robust feature representations. It aligns speech sequences with their corresponding phonemes during training, eliminating the need for explicit frame-level annotations.

In addition, we introduce dynamic curriculum learning, which dynamically organizes negative samples based on difficulty, determined by the similarity distance of the anchor and negative phoneme in PCL. This phonetic approach further enhances DSR performance by effectively distinguishing between similar-sounding phonemes, a critical factor in dysarthric speech.


Evaluated on the UASpeech dataset, a representative DSR benchmark, our method shows substantial improvements over baseline models. Specifically, in the lowest intelligibility group, DyPCL reduces the word error rate (WER) from 58.49\% to 49.45\%, while overall WER across all dysarthria groups drops from 25.97\% to 20.23\%. Extensive analysis and ablation studies highlight the robustness of the proposed strategies.

\section{Related Work}



\paragraph{Dysarthric Speech Recognition} Prior studies have primarily utilized data augmentation and speaker-adaptive training to address DSR challenges. Augmentation methods like speed and temporal perturbation \cite{PranantaH0S22, geng2022speaker} simulate dysarthric speech characteristics, while adversarial training \cite{jiao2018simulating, huang2022towards, jin2023adversarial, jin2023personalized, wang2024enhancing} and diffusion models \cite{wang2023duta} synthesize dysarthric speech. Speaker-adaptive training helps models handle speaker variability through features like Learning Hidden Unit Contributions \cite{yu2018development, geng2022fly, geng2023use}, x-vector \cite{baskar2022speaker}, and Acoustic-to-Articulatory inversion models \cite{hu2019cuhk, liu2020exploiting, hu2022exploiting, hu2024self, Hsieh2024, linexploring, hu2023exploring}. However, these methods require additional datasets and external models, which increases computational complexity. We simplify the DSR process by using only the UASpeech dataset and a single ASR model, eliminating the need for external resources.

\paragraph{Contrastive Learning} Contrastive learning in speech recognition has proven effective in various atypical scenarios, such as noisy environments \cite{wang2022wav2vec, zhu2023robust} and accented speech datasets \cite{han2021supervised, scala}, due to its ability to learn robust speech representations by reducing speech embedding variability. For DSR, contrastive learning has been applied to reduce the distance between healthy utterance and dysarthric utterance representations \cite{wu2021sequential, wang2024enhancing} using the word or utterance fragments. 
To consider subtle segments during training, phoneme-level contrastive learning has been suggested for accented speech recognition \cite{scala}, but it has yet to be explored in DSR.

To perform phoneme-level contrastive learning, sophisticated phoneme alignment for a given user utterance is required. Previous research explored various alignment methods \cite{rousso2024, gorman2011prosodylab, montreal} to match given phonemes to audio frames closely. Notably, using CTC forced alignment has shown great alignment accuracy \cite{huang2024less, zhao2024advancing}.
This work proposes phoneme-level dynamic CTC alignment that dynamically adapts during optimizing contrastive learning. In addition, considering the importance of hard negative sampling for contrastive learning \cite{Robinson2021Contrastive, kalantidis2020hard}, we dynamically select negative samples across varying difficulty groups and phoneme distance levels, integrating with the curriculum learning process; thus, the model can learn invariant representations.




\begin{figure}[t!]
    \centering
    \includegraphics[width=1\linewidth]{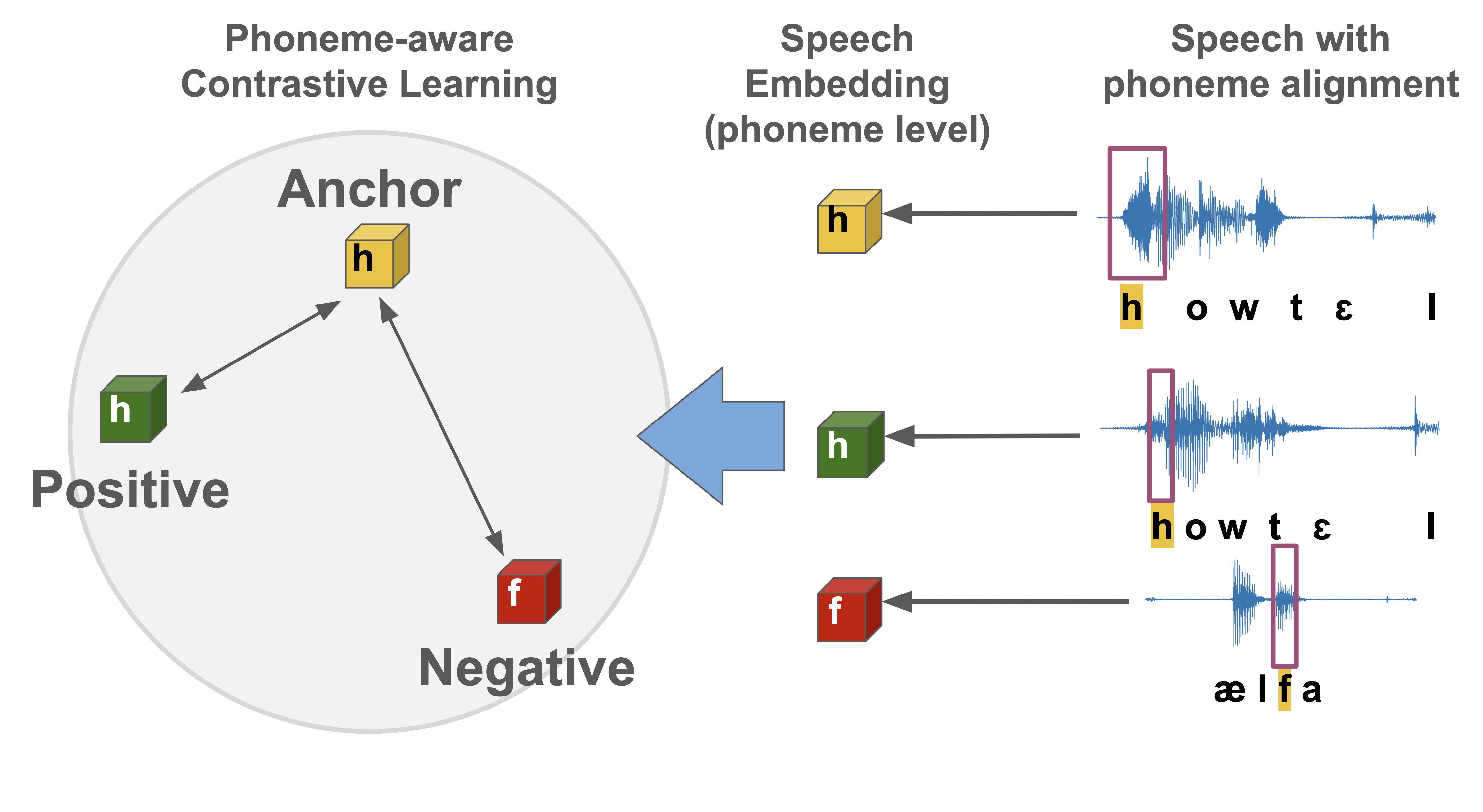}
    \caption{Phoneme-level Contrastive Learning. The phoneme-aligned speech segment corresponding to the phoneme "h" in the word "HOTEL" is used as both the anchor and the positive sample, while the phoneme "f" from the word "ALPHA" serves as the negative sample.}
    \label{fig:overview}
\end{figure}





\section{DyPCL}

\begin{figure*}[t]
    \centering
    \includegraphics[width=0.95\linewidth]{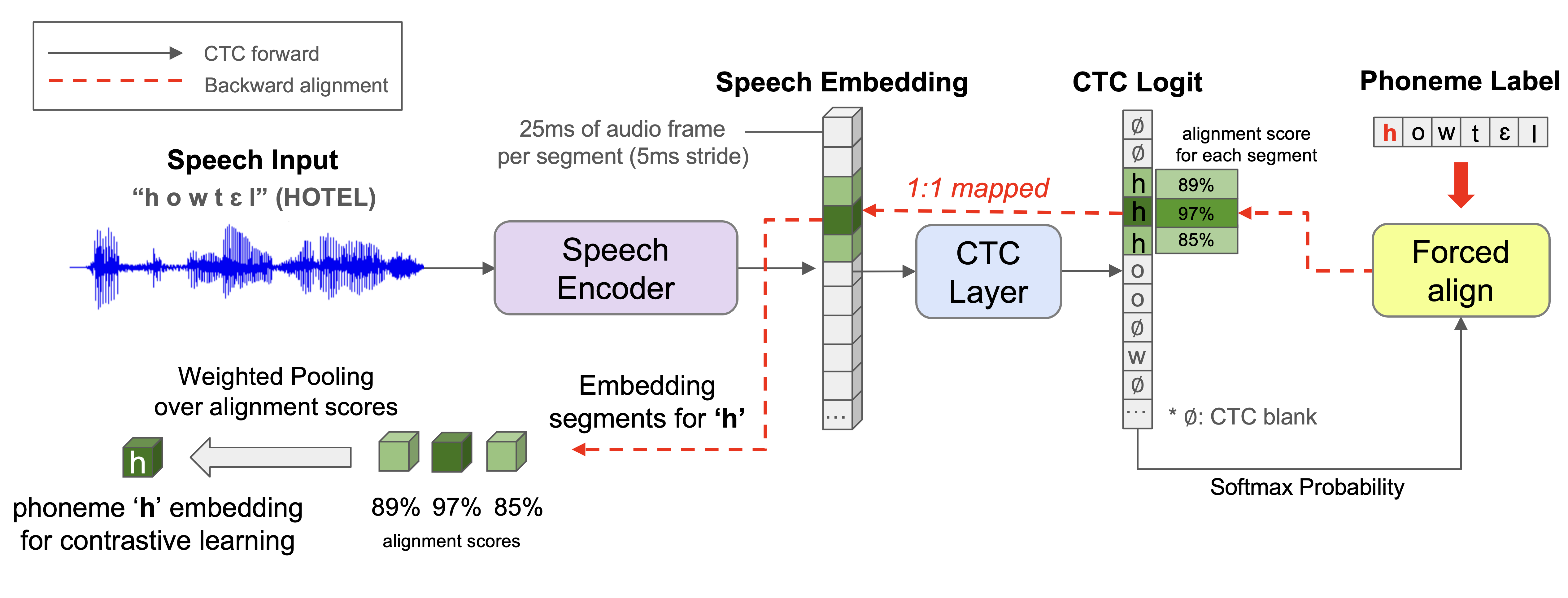}
    \caption{Phoneme embedding extraction for the target phoneme "h" in the word "HOTEL" using \textbf{Dynamic CTC Alignment} for phoneme-level contrastive learning.}
    \label{fig:align}
\end{figure*}

\subsection{Phoneme-level Contrastive Learning}
To conduct DyPCL, we leverage phoneme-level contrastive learning (PCL) with CTC loss in a multitask learning framework. PCL is a training strategy designed to learn phoneme-level representations for speech recognition through contrastive learning (Figure \ref{fig:overview}). PCL effectively clusters and separates targeted phoneme embeddings by focusing on audio segments corresponding to phonemes. This approach is particularly robust for tasks involving single-word audio, where even a minor phoneme error can significantly impact intelligibility. Further, PCL benefits CTC models that use the same phonemes as output units by enhancing the model's ability to distinguish subtle phonetic variations, thereby improving recognition accuracy.


We design two training stages: (1) CTC training and (2) combined CTC and PCL training for contrastive learning. The first stage, CTC training, targets phoneme recognition based on given speech and labels, following procedures in general speech recognition. In the second stage, CTC/PCL training integrates phoneme-level contrastive learning specifically for dysarthric speech while maintaining CTC training for phoneme recognition.

We pair anchor, positive, and negative samples to construct the contrastive learning dataset from UASpecch \cite{uaspeech-data}. We decompose words into phonemes using the phonemizer tool \cite{epitran}. Anchor samples are exclusively selected from the control (C) group, healthy speech, to serve as a reference for correct pronunciation. Positive samples are the same word and phoneme as the anchor but taken from dysarthric groups (H, M, L, VL) (refer to \ref{sec:data}). Negative samples comprise different words and phonemes from the anchor and positive, also drawn from dysarthric groups to introduce challenging contrasts. For the healthy-speech-only trainset (B2-Control of UASpeech), positive samples are selected from different speakers within the same group. 

This process yields roughly 48.65 billion triplet pairs. We use stratified sampling to balance training cost and efficiency, limiting each anchor to a maximum of five positive samples and each anchor-positive pair to five negative samples. Consequently, 1.18 million triplet pairs are created for contrastive learning. During training, we randomly sampled 200,000 triplet pairs.

The triplet loss works by ensuring that the embedding of an anchor sample \( a \) is closer to a positive sample \( p \) than to a negative sample \( n \), by at least a given margin. The triplet loss, $L_{\text{triplet}}(a, p, n)$, with margin $m$ is defined as:

\begin{small}
\begin{equation}
\begin{split}
\max \Big( 0,  \, \| f(\mathbf{a}) - f(\mathbf{p}) \|_2^2 - \| f(\mathbf{a}) - f(\mathbf{n}) \|_2^2 + {m} \Big)
\end{split}
\label{eq:triplet}
\end{equation}
\end{small}

Here, \( f(x) \) represents the speech embedding for phoneme \( x \) obtained from the speech encoder, where \( f(\mathbf{a}) \) is the embedding of the anchor, \( f(\mathbf{p}) \) is the embedding of the positive sample, and \( f(\mathbf{n}) \) is the embedding of the negative sample. The squared Euclidean distances \( \| \cdot \|_2^2 \) measure how far apart these embeddings are. The margin ensures the anchor is closer to the positive than the negative by a certain threshold, encouraging the model to learn distinct representations for different phonemes. 
Our total multitask loss function for CTC/PCL training, $L_{\text{total}}$, is defined as follows:
\begin{equation}
\begin{split}
 \Big(\frac{1}{3}\sum_i^{(A,P,N)} L_{\text{CTC}}(i) \Big) + \lambda \cdot L_{\text{triplet}}(a, p, n)
\end{split}
\label{eq:total}
\end{equation}
where \( A \), \( P \), and \( N \) denote the anchor, positive, and negative audio samples, respectively, and \( a \), \( p \), and \( n \) are the corresponding anchor, positive, and negative phonemes. In this work, we set \( \lambda = 0.5 \). The following subsections outline the dynamic components of PCL, forming the basis of DyPCL.


\begin{figure*}
    \centering
    \includegraphics[width=0.95\linewidth]{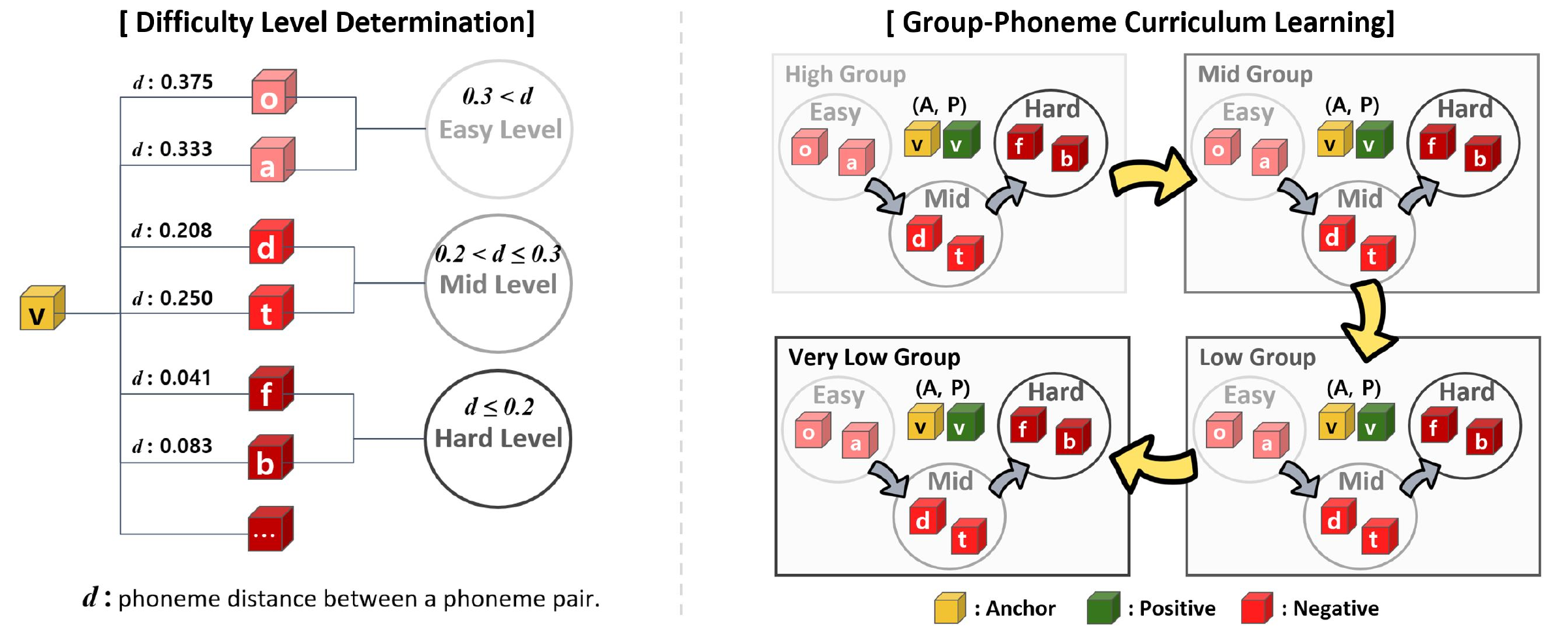}
    \caption{The illustration of difficulty level determination in three (easy, mid, hard) levels by phoneme distance measurement (Left) and the group-phoneme (GP) curriculum learning (Right). The figure shows an example where the anchor and positive samples are "v". }
    \label{fig:curriculum}
\end{figure*}

\subsection{Dynamic CTC Alignment}
\label{sec:dyctc}

To perform PCL, we need a phoneme-level alignment for each audio sample. This requires accurately mapping each phoneme to its corresponding speech embedding.
\citet{scala} demonstrate that phoneme-level contrastive learning can improve ASR accuracy in accented speech. They utilize an HMM-DNN acoustic model for forced alignment using the Kaldi toolkit \cite{kaldi}. Although the authors acknowledge that the alignment may not be perfect, their results show substantial improvements with phoneme-level contrastive learning.

In contrast, our research focuses on dysarthric speech recognition, where pre-trained forced alignment models often struggle to accurately align speech with labels, particularly for groups with low intelligibility. This misalignment arises because these models are typically trained on standard speech datasets, while the acoustic properties of dysarthric speech vary significantly from typical speech. Such discrepancies in alignment can critically impair phoneme-level contrastive learning.

Forced alignment is conventionally done by timestamping \cite{rousso2024}, which provides phoneme boundaries about audio frames. However, for our purposes, we need a method that pinpoints the corresponding speech embedding for each phoneme within a CTC model; thus, a direct solution is required to eliminate unnecessary errors. Figure \ref{fig:align} presents an overview of the proposed dynamic CTC alignment. Drawing from CTC forced alignment\footnote{\url{https://pytorch.org/audio/main/generated/torchaudio.functional.forced_align.html}}, we directly extract speech embeddings for specific phonemes. CTC forced alignment maps audio to transcription by predicting the most likely alignment between speech frames and text based on CTC logits, handling timing variations.

The output provides alignment scores for each CTC logit. Given the speech embeddings shaped $[\textit{embedding size}, \textit{sequence length}]$ and CTC logits shaped $[\textit{CTC vocabulary size}, \textit{sequence length}]$, we can map the alignment scores to the corresponding speech embeddings for each phoneme, as illustrated by the red lines (backward alignment) in Figure \ref{fig:align}. As the speech encoder (wav2vec2.0 \cite{baevski2020wav2vec} and its variants) generates one embedding token per 25 ms of audio, multiple indices in both the embedding and logits can map to a single phoneme. We generate a single phoneme representation using weighted pooling with alignment scores as weights. The weighted pooled phoneme embeddings for each anchor (\( f(\mathbf{a}) \)), positive (\( f(\mathbf{p}) \)), and negative (\( f(\mathbf{n}) \)) samples will be used in the triplet loss, as defined in Equation \ref{eq:triplet}.
The speech encoder and CTC layer will be updated during CTC/PCL training, leading to more accurate alignments, and these improved alignments will further enhance the training process.


\subsection{Dynamic Curriculum Learning}
\label{sec:curriculum}
Negative sampling in contrastive learning is crucial, as selecting hard negatives can significantly enhance model performance \cite{Robinson2021Contrastive, kalantidis2020hard, srinidhi2021improving}. Each phoneme is treated as an anchor, positive, or negative sample in PCL.
Anchors are selected from the control group (C), serving as the reference, while positives are chosen from the same word utterance as the anchor but from the dysarthric group (H, M, L and VL. Refer to Table \ref{tab:uaspeech}). Negatives are randomly sampled from other phonemes within the dysarthric group, which do not directly relate to the anchor or positive.

In the DyPCL framework, we dynamically select negative samples using a curriculum learning approach. This dynamic selection strategy gradually increases the difficulty of negative samples, fostering a more robust learning process.
The difficulty is determined by phoneme distance, which is measured using the PanPhon tool \cite{panphon} \footnote{\url{https://github.com/dmort27/panphon}}. 
The \textit{hamming feature edit distance} is used to calculate phoneme distance, with equal weighting across all 21 articulatory features that define each phoneme. The phoneme distance reflects how similar or different phonemes sound phonetically, ranging from 0.0416 (most similar) to 0.583 (most different).
Figure \ref{fig:heatmap} in Appendix \ref{sec:appendix} shows the phoneme distance matrix in a heat map.
By learning to distinguish similar-sounding phonemes in the embedding space through DyPCL, the model can further improve recognition accuracy for given phonemes.

We differentiate the curriculum by varying the phoneme distance of negative samples in multiple ways. As shown on the left side of Figure \ref{fig:curriculum}, we categorize the difficulty into three levels: easy ($d > 0.3$), medium ($0.2 < d \leq 0.3$), and hard ($d \leq 0.2$), where $d$ represents the phoneme distance between an anchor and a negative sample. Further difficulty variations are discussed in Section \ref{sec:difficulty}.

To further optimize the effectiveness of our phoneme distance-based curriculum (\textbf{P}), we incorporate a group-level curriculum (\textbf{G}) as suggested in \citet{hsieh2024dysarthric}. This method trains the DSR model progressively, following an intelligibility group order from H to M, L, and VL, which has improved DSR accuracy compared to non-ordered training. We enhance the effectiveness of the curriculum by combining both P and G strategies.

The resulting \textbf{GP} (group first, then phoneme distance) curriculum first organizes the groups in the H, M, L, and VL order and then applies the P strategy within each group. The GP curriculum comprises 12 levels (4 groups $\times$ 3 phoneme difficulty levels), as illustrated on the right side of Figure \ref{fig:curriculum}. The \textbf{PG} (phoneme distance first, then group) curriculum is also evaluated in Section \ref{sec:result.main}.
All curricula were designed to train on 200,000 triplet pairs per epoch.





\section{Experiment Setup}






\subsection{Model \& Training}

In our experiments, we utilize a CTC head with several pretrained speech encoders, including Wav2Vec2.0\footnote{\url{https://huggingface.co/facebook/wav2vec2-large-960h}}\cite{baevski2020wav2vec}, HuBert\footnote{\url{https://huggingface.co/facebook/hubert-large-ls960-ft}}\cite{hsu2021hubert}, and WavLM\footnote{\url{https://huggingface.co/microsoft/wavlm-large}}\cite{chen2022wavlm}, each with 315M parameters and a CTC head of 44K parameters.
All speech encoders have an embedding size of 1024. We extract phoneme-level speech embeddings \( f(x) \) sized [1024, 1] via dynamic CTC alignment for PCL.

We first trained the model using only CTC loss in the initial stage, followed by combined CTC/PCL training. This two-stage approach helped refine the dynamic CTC alignment for optimal performance in PCL training. The model was optimized using the AdamW \cite{loshchilov2017decoupled} algorithm with parameters \( (\beta_1, \beta_2) = (0.9, 0.99) \), a learning rate of \( 3 \times 10^{-4} \), weight decay of \( 1 \times 10^{-5} \), and batch sizes of 128 for CTC training and 64 for CTC/PCL training. We trained the models on 8 NVIDIA A100 GPUs, utilizing a linear learning rate scheduler and selecting the best model based on the lowest overall WER on the validation set. The CTC training ran for 100 epochs (10 hours), and the CTC/PCL stage ran for 5 epochs (20 hours).

\subsection{Dataset \& Preprocessing}
\label{sec:data}
The recent release of UASpeech \cite{uaspeech-data} includes 13 speakers in a control group (denoted as C) and 15 speakers with dysarthria, categorized into intelligibility levels: High (H), Mid (M), Low (L), and Very Low (VL). For more details, please refer to Table \ref{tab:uaspeech} in Appendix \ref{sec:appendix}.

We follow prior works for training split (\textbf{TRAIN}) \cite{hu2022exploiting, geng2023use, hu2024self, hsieh2024dysarthric} which has all audio from B1, B3 and B2 of control group (B2-Control). For test splits, we use two sets: B2 of all dysarthria groups (\textbf{TEST}) \cite{geng2023use, hu2024self, hsieh2024dysarthric}, and only the common words (CW) of dysarthria group excluding uncommon words (UW) (\textbf{CTEST}) \cite{bhat2022improved, proto}. We randomly sampled 10\% of the TRAIN set for validation split. Every audio recording is from microphone 5 (M5) in UASpeech.

\begin{table*}[]
\centering
\scalebox{
0.9}{%
\begin{tabular}{l|c|c|c||ccccc}
\toprule
\multicolumn{4}{c||}{Configuration} & \multicolumn{5}{c}{UASpeech WER (\%)} \\ 
\midrule
Speech Encoder & Loss Function & CL Target & Neg. Sampling & H & M & L & VL & ALL \\ 
\midrule
\midrule
Wav2Vec2.0 & CTC & - & - & 6.54 & 21.55 & 31.89 & 61.04 & 29.15 \\
WavLM & CTC & - & - & 5.32 & 20.10 & 26.43 & 58.94 & 26.80 \\
HuBERT \checkmark& CTC & - & - & \textbf{4.33} & \textbf{19.32} & \textbf{25.32} & \textbf{58.49} & \textbf{25.97} \\
\midrule
\multirow{6}{*}{HuBERT \checkmark} & CTC+CL & word    & R  & 5.84 & 18.63 & 24.40 & 58.50 & 26.15 \\
                        & CTC+CL & phoneme & R  & \textbf{3.12} & \textbf{15.21} & \textbf{21.16} & \textbf{53.72} & \textbf{22.64} \\
\cmidrule{2-9}
                        & CTC+CL & phoneme & G  & 3.24 & 15.43 & 20.81 & 50.77 & 21.87 \\
                        & CTC+CL & phoneme & P  & 2.75 & 14.56 & 18.45 & 51.26 & 21.19 \\
                        & CTC+CL & phoneme & PG & \textbf{2.73} & 13.21 & 18.20 & 50.21 & 20.58 \\
                        & CTC+CL & phoneme & GP & 2.77 & \textbf{12.98} & \textbf{17.60} & \textbf{49.45} & \textbf{20.23} \\
\bottomrule
\end{tabular}%
}
\caption{WER on UASpeech \textbf{TEST} set with different configurations: Speech Encoder, Loss function, Contrastive Learning (CL) target, and Negative sampling method. \textbf{R} in negative sampling represent random sampling and \textbf{G}, \textbf{P}, \textbf{PG} and \textbf{GP} represent corresponding curriculum strategy described in Section \ref{sec:curriculum}.
"ALL" denotes the average WER across four groups, weighted by the number of speakers in each group.}
\label{tab:main}
\end{table*}

\section{Result \& Analysis}
\subsection{Main Result}
\label{sec:result.main}

In Table \ref{tab:main}, we first evaluate the performance of three speech encoders—Wav2Vec2.0, WavLM, and HuBERT—using CTC training. As indicated in several studies \cite{proto, hu2024self}, the HuBERT model achieves the best WER across all speaker groups and the overall average. Consequently, we conducted further experiments using the HuBERT model.

To assess the impact of phoneme-level contrastive learning, we trained the HuBERT-CTC model using word- and phoneme-level contrastive learning. Since the UASpeech dataset consists of isolated word recordings, word-level alignment was not required for word-level contrastive learning. Our results indicate that word-level contrastive learning underperformed, even falling short of the baseline CTC model. In contrast, phoneme-level contrastive learning significantly reduced WER across all speaker groups and the overall average. Notably, the VL group showed a marked improvement, with WER decreasing from \textbf{58.49\%} using CTC alone to \textbf{53.72\%} with PCL.

Then, we evaluate the negative sampling strategies within PCL. 
All curriculums (G, P, PG, GP) show improvements compared to the PCL model with random negative sampling (R). When comparing group-level (G) and phoneme-level (P) curricula, we found that P achieved better overall performance (21.18\% vs. 21.87\%), though G performed better in the VL group (50.77\% vs. 51.26\%). We attribute this to the fact that the VL group is trained last in the G curriculum.

When combining both strategies, \textbf{PG} and \textbf{GP} yielded substantial improvements over the individual methods. In particular, the \textbf{GP} curriculum achieved the best overall WER of \textbf{20.23\%} and \textbf{49.45\%}in VL, demonstrating that curriculum learning with a sophisticated difficulty progression can further enhance DSR performance.

\begin{table}[]
\centering
\resizebox{\columnwidth}{!}{%
\begin{tabular}{l|ccccc}
\toprule
\multirow{2}{*}{Phoneme alignment method}                      & \multicolumn{5}{c}{UASpeech WER(\%)}                                               \\ \cline{2-6} 
                                                       & H             & M              & L              & VL             & ALL            \\ \midrule\midrule
CTC forced align (timestamp)                          & 3.67          & 19.21          & 26.49          & 58.71          & 26.02           \\
\multicolumn{1}{c|}{CTC forced align (logit level)}    & 3.65          & 18.32          & 26.91          & 57.43          & 25.58          \\
\multicolumn{1}{c|}{\textbf{Dynamic CTC alignment}} & \textbf{3.12} & \textbf{15.21} & \textbf{21.16} & \textbf{53.72} & \textbf{22.64} \\ \bottomrule
\end{tabular}%
}
\caption{Effect of different alignment methods for PCL. Curriculum learning is not applied in this evaluation (random sampling is used) to isolate the impact of the alignment methods.}
\label{tab:alignment}
\end{table}

\begin{figure*}[ht!]
    \centering
    \includegraphics[width=0.90\linewidth]{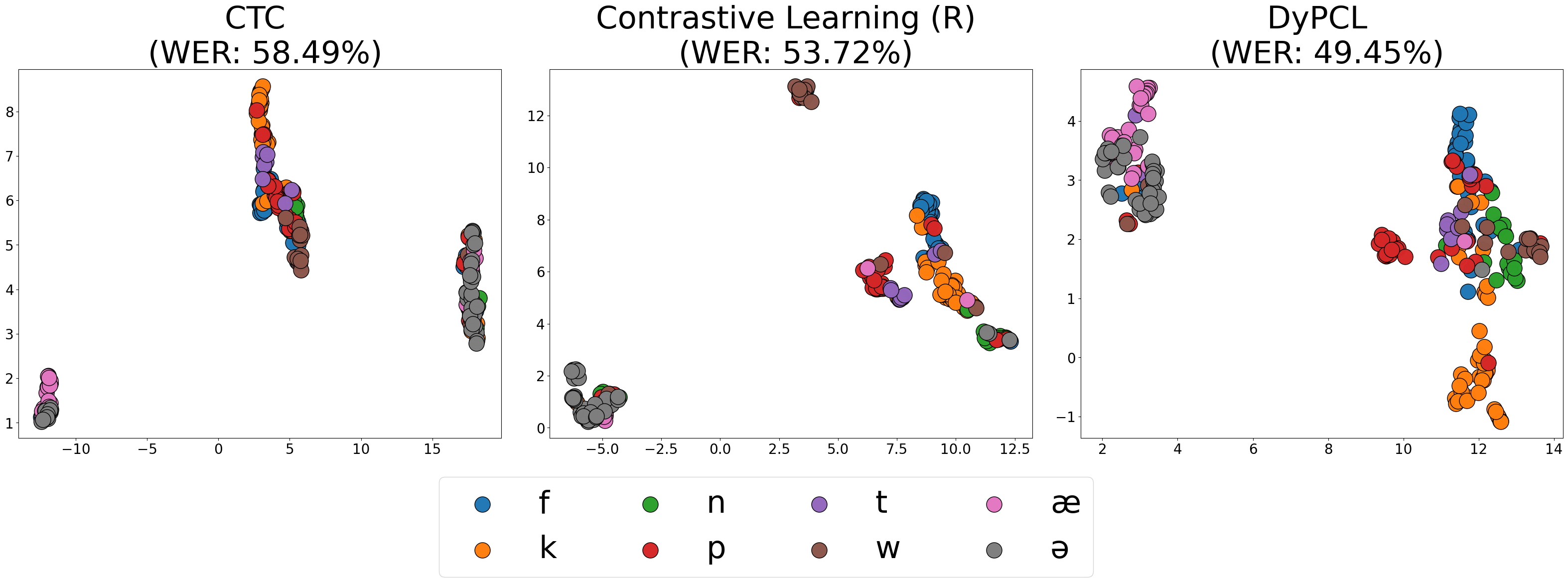}
    \caption{UMAP Visualization of Phoneme embeddings on \textbf{TEST} set (\textbf{VL} Group only): Phoneme embeddings, extracted via forced CTC alignment (Figure \ref{fig:align}), are shown for three models: CTC, Contrastive learning with random sampling (\textbf{R}), and DyPCL with group-phoneme level curriculum (\textbf{GP}). Points are color-coded by phoneme, illustrating how each model clusters and separates them. For each phoneme, up to 100 embeddings were displayed.}
    \label{fig:UMAP}
\end{figure*}

\subsection{Effect of Dynamic CTC Alignment}

As discussed in Section \ref{sec:dyctc}, conventional phoneme alignment models struggle to accurately align phonemes in dysarthric speech, particularly for speakers with low intelligibility. This misalignment can severely impact the effectiveness of PCL, as it relies on extracting precise phoneme embeddings. To address this, we evaluate the effectiveness of dynamic CTC alignment by comparing it to conventional alignment methods.

For a fair comparison, we use the HuBERT-CTC model trained on the \textbf{TRAIN} set as a baseline for CTC forced alignment (Table \ref{tab:main}). Phoneme alignment can be implemented in two primary ways: (1) using the timestamps of target phonemes in the audio to extract phoneme embeddings by calculating their corresponding embedding indices based on a 25ms window with a 5ms stride \cite{baevski2020wav2vec}, and (2) applying backward alignment on CTC logits, as proposed in the dynamic CTC alignment. The key difference between the second approach and dynamic CTC alignment is that the alignment model in the former is not updated during training.

In Table \ref{tab:alignment}, the use of alignment with timestamps yielded underwhelming results, showing only marginal improvements for the H and M groups compared to the HuBERT-CTC model results in Table \ref{tab:main}. This result can be attributed to incorrect alignments and potential conversion errors between timestamps and phoneme embeddings. When alignment was applied at the logit level, we achieved overall improvements, though there was some degradation in the VL group. In contrast, dynamic CTC alignment substantially improved WER across all groups. These gains are attributed to the model being optimized with both CTC and PCL losses, which enhance alignment accuracy and, in turn, lead to better DSR accuracy.




\subsection{Analysis on Curriculum Difficulty Levels}
\label{sec:difficulty}

\begin{figure}[t]
    \centering
    \includegraphics[width=0.95\linewidth]{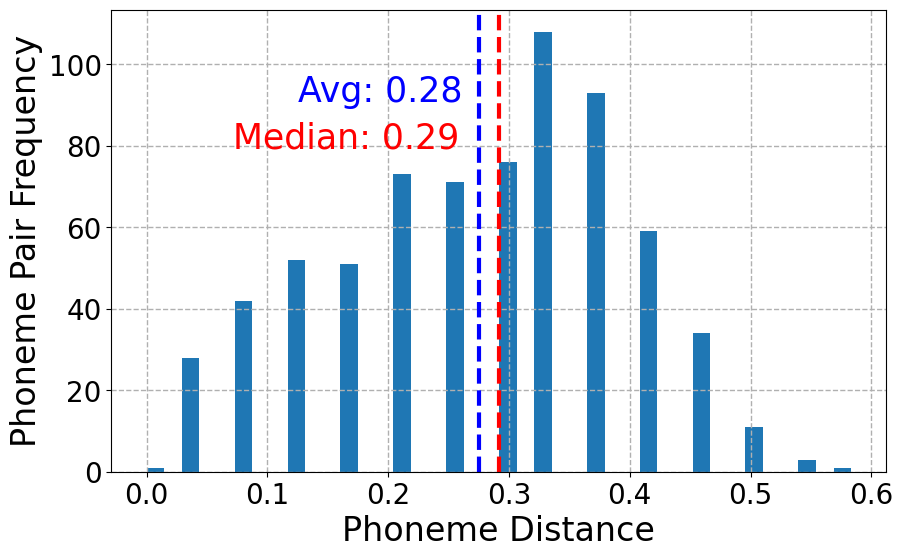}
    \caption{Distribution of Phoneme distance over phoneme pairs.}
    \label{fig:freq}
\end{figure}

In Figure \ref{fig:freq}, the average and median phoneme distances are 0.28 and 0.29, respectively. Based on this, we initially set the threshold at 0.3 to divide the difficulty into two levels: easy and hard (2 LV). We then refined the division by adding a mid-difficulty level at 0.2, creating three levels: easy, mid, and hard (3 LV). Furthermore, we explored finer granularity by dividing the phoneme distance range into 0.1 intervals, resulting in six difficulty levels (6 LV).

\begin{table}[]
\centering
\resizebox{\columnwidth}{!}{%
\begin{tabular}{l|ccccc}
\toprule
\multirow{2}{*}{\begin{tabular}[c]{@{}l@{}}Phoneme distance ranges\end{tabular}} & \multicolumn{5}{c}{UASpeech WER(\%)}                                     \\ \cline{2-6} 
                                                                                   & H             & M              & L             & VL             & ALL   \\ \midrule
(2 LV) $0 < H \leq 0.3 < E \leq 0.583$                                             & \textbf{2.75} & 13.25          & 17.94         & 50.43          & 20.6  \\
(3 LV) $0 < H \leq 0.2 < M \leq 0.3 < E \leq 0.583$                                & 2.77          & \textbf{12.98} & \textbf{17.6} & \textbf{49.45} & \textbf{20.23} \\
(6 LV) divide every $0.1$ distance                                                 & 2.94          & 13.70          & 18.03         & 51.43          & 21.04 \\ \bottomrule
\end{tabular}%
}
\caption{Difficulty levels for phoneme distance curriculum. WER is evaluated on \textbf{TEST}.}
\label{tab:level}
\end{table}


Table \ref{tab:level} illustrates how difficulty levels in dynamic curriculum learning were established by segmenting the range of phoneme distances. Our results show that dividing the phoneme distance into three levels (easy, mid, and hard) yielded the best performance for our DyPCL. Although a two-level division produced comparable results, the three-level split outperformed it overall. However, the six-level division, which introduces extremely easy ($d \leq 0.1$) and extremely hard ($0.5 \leq d$) negative pairs, led to suboptimal results. This comparison highlights the trade-offs between granularity and performance, as the narrow distance intervals often resulted in some phonemes lacking suitable negative pairs, which adversely affected performance in our evaluations.
Figure \ref{fig:bars} in Appendix \ref{sec:appendix} shows the distribution of phoneme pairs across the three difficulty levels (3 LV) for each phoneme.

\subsection{Phoneme Embedding Discrimination and Clustering}
Figure \ref{fig:UMAP} presents the Uniform Manifold Approximation and Projection \cite{mcinnes2018umap-software} (UMAP) visualization of phoneme embeddings for the \textbf{TEST} VL group across three models. The distribution of phoneme embeddings illustrates how well each model distinguishes and clusters the phonemes.
In the CTC model, phoneme embeddings are not well-separated and form small, unclear clusters. The contrastive learning model shows more distinct clustering, although some phonemes still appear ambiguously grouped. In contrast, the DyPCL model demonstrates a clear and decisive separation of phonemes. Notably, even very similar-sounding phonemes, such as "\textbf{\ae}"  and "\textbf{\rotatebox[origin=c]{180}{e}}" (marked in pink and grey, respectively) with a phoneme distance of 0.125, are well-separated in DyPCL. This improvement is attributed to the informative curriculum learning strategy, which progressively trains the model to better distinguish similar phonemes.

\subsection{Comparison with Benchmarks}


Table \ref{tab:bench} presents a comparison of the performance of our model against state-of-the-art (SOTA) methods on the \textbf{CTEST} set. \citet{Bhat2022} employs a two-stage augmentation approach. The second and third methods come from \citet{proto}, which was the first to introduce contrastive learning for UASpeech. In their study, the "Speaker Dependent (SD) w/ finetune" method focuses on fine-tuning speaker-specific word prototypes, while the "Speaker Independent (SI)" method works across both word- and speaker-level instances to improve generalization. The SD approach improved the WER to 13.49\%, while the SI method further reduced it to 12.09\%. 
In contrast, our model, DyPCL (GP), achieved the lowest WER at severity levels, significantly reducing the overall WER to 10.34\%, outperforming all previous models.

Table \ref{tab:bench} also compares our model, DyPCL (GP), against other methods on the \textbf{TEST} set. For a fair comparison, the results from previous works are reported without data augmentation (DA), focusing on the core contributions of each method. \citet{wang2023hyper} used hyperparameter adaptation to handle speaker differences, achieving an overall WER of 30.49\%. \citet{Hsieh2024} applied curriculum learning, training progressively from high to low intelligibility groups with their proposed re-grouping method, but still reported a relatively high WER of 7.99\% for the easiest group (H), highlighting challenges in early-stage learning.
Both \citet{hu2023exploring} and \citet{hu2024self} used speaker-adaptive training, incorporating speaker-specific articulatory and acoustic features, achieving WERs of 24.62\% and 22.62\%, respectively. \citet{geng2023use} integrated severity information and system combination, and without DA, reported an overall WER of 24.57\%.


Our model, DyPCL (GP), not only achieved the lowest overall WER of 20.7\%, outperforming all previous methods but also demonstrated significant improvements for the low intelligibility groups. With WERs of 17.6\% for the L group and 49.45\% for the VL group, it maintained strong performance across all dysarthria severity levels, achieving a WER of just 2.77\% for the High (H) intelligibility group. This level of robustness highlights the reliability of its performance.

In Table \ref{tab:bench}, "ALL*" represents the recalculated unweighted average WER across the four dysarthria groups, ensuring consistency and fairness in our comparison process. It is important to note that variations in reported WERs in different papers may arise from additional factors, such as the inclusion of control groups, which we have taken into account.

\begin{table}[t]
\centering
\resizebox{\columnwidth}{!}{%
\begin{tabular}{l|ccccc}
\toprule
\multirow{2}{*}{Model} & \multicolumn{5}{c}{UASpeech WER(\%)}                                             \\ \cline{2-6} 
                                & H             & M             & L             & VL             & ALL*            \\ \midrule\midrule
\multicolumn{6}{c}{\textbf{CTEST}} \\ \midrule
\citet{Bhat2022}                             & 6.40           & 14.6          & 18.9          & 61.50           & 25.35           \\
\citet{proto} (SD w/ finetune)          & 5.12          & 4.89          & 6.27          & 37.67          & 13.49          \\
\citet{proto} (SI)                      & 2.35          & 6.01          & 7.91          & 32.11          & 12.09          \\
DyPCL (GP)                            & \textbf{1.09} & \textbf{3.94} & \textbf{5.02} & \textbf{31.33} & \textbf{10.34} \\ \midrule\midrule
\multicolumn{6}{c}{\textbf{TEST}} \\ \midrule
\citet{wang2023hyper}   & 5.22                 & 21.35                & 33.37                & 62.04                & 30.49                \\
\citet{Hsieh2024} & 7.99                 & 16.12                & 22.28                & 52.15                & 24.64                \\
\citet{hu2023exploring}      & 6.32                  & 14.04                & 25.03                & 53.12                 & 24.62                \\
\citet{geng2023use} (w/o DA)      & 2.91                  & 12.10                & 23.91                & 59.38                 & 24.57                \\
\citet{hu2024self}      & 4.20                  & \textbf{12.06}              & 23.51                & 50.7                 & 22.62                \\
DyPCL (GP)             & \textbf{2.77}        & 12.98       & \textbf{17.6}        & \textbf{49.45}       & \textbf{20.7}        \\ \bottomrule
\end{tabular}%
}
\caption{WER comparison on the \textbf{CTEST} and \textbf{TEST}, showing the performance of DyPCL (GP) against previous studies. *: unweighted average over groups.}
\label{tab:bench}
\end{table}

\section{Conclusion}

This paper introduced the Dynamic Phoneme-level Contrastive Learning (DyPCL) framework to improve dysarthric speech recognition. DyPCL effectively tackles phoneme alignment challenges and accounts for phonetic difficulty through dynamic CTC alignment and curriculum learning.
Our experiments on the UASpeech dataset demonstrated the effectiveness of DyPCL, which reduced the WER from 58.49\% to 49.45\% in the Very Low (VL) intelligibility group and the overall WER across all dysarthria groups from 25.97\% to 20.23\%. These results underscore DyPCL’s capability to capture subtle phonetic variations, significantly enhancing speech recognition accuracy across all levels of dysarthria severity




\section*{Limitations}
While DyPCL has shown strong performance in recognizing dysarthric speech, its reliance on paired data, such as in the UASpeech dataset, where each dysarthric speech sample is paired with a corresponding control group utterance, suggests that there may be opportunities to further generalize the model. This pairing provides valuable reference points for contrastive learning, but by focusing on phoneme embeddings rather than specific word pairs, future research could explore the model’s applicability in scenarios where such paired data is unavailable. This shift could potentially broaden DyPCL's utility in more diverse environments where only unpaired or less structured data is available.

Moreover, we did not employ data augmentation (DA) techniques in this study to ensure a clear evaluation of DyPCL’s core contributions. However, given that previous research indicates the positive impact of DA on dysarthric speech recognition, combining DyPCL with DA strategies could yield further improvements. Future work will explore these possibilities to enhance performance and generalizability.

\section*{Acknowledgments}

This work was partly supported by Institute of Information \& communications Technology Planning \& Evaluation (IITP) grant funded by the Korea government(MSIT) (No.RS-2019-II191906, Artificial Intelligence Graduate School Program(POSTECH)) (5\%) and was supported by the MSIT(Ministry of Science and ICT), Korea, under the ITRC(Information Technology Research Center) support program(IITP-2024-RS-2024-00437866) supervised by the IITP(Institute for Information \& Communications Technology Planning \& Evaluation) and was supported by Smart HealthCare Program(www.kipot.or.kr) funded by the Korean National Police Agency(KNPA, Korea) [Project Name: Development of an Intelligent Big Data Integrated Platform for Police Officers’ Personalized Healthcare / Project Number: 220222M01] 

\bibliography{acl_latex}

\newpage
\appendix

\section{Appendix}
\label{sec:appendix}

\begin{table}[h!]
\centering
\scalebox{0.69}{
\begin{tabular}{cccc}
\toprule
Dysarthria Group & Speaker ID &\begin{tabular}[c]{@{}c@{}}Speech\\ Intelligibility (\%) \end{tabular}\ &  \begin{tabular}[c]{@{}c@{}}\# Uttr.\\ (CW/UW)\end{tabular} \\ 
\midrule
\multirow{5}{*}{High \textbf{(H)}} & F05 & 95 & 465/300 \\
                      & M08 & 93 & 465/300 \\
                      & M10 & 93 & 465/300 \\
                      & M14 & 90 & 465/300 \\
                      & M09 & 86 & 465/300 \\ 
\midrule
\multirow{3}{*}{Mid \textbf{(M)}}  & F04 & 62 & \textbf{461}/\textbf{289} \\
                      & M11 & 62 & 465/300 \\
                      & M05 & 58 & 465/300 \\ 
\midrule
\multirow{3}{*}{Low \textbf{(L)}}  & M16 & 43 & 465/300 \\
                      & F02 & 29 & 465/300 \\
                      & M07 & 28 & 465/300 \\ 
\midrule
\multirow{4}{*}{Very Low \textbf{(VL)}} & M01 & 15 & 465/300 \\
                          & M12 & 7  & 465/300 \\
                          & F03 & 6  & \textbf{451}/300 \\
                          & M04 & 2  & 465/300 \\ 
\bottomrule
\end{tabular}}
\caption{Speech intelligibility levels and number of utterances for dysarthric speakers in the UASpeech \cite{uaspeech-data} dataset, ordered by intelligibility. The "CW/UW" denotes the number of utterances for common words (CW) and uncommon words (UW). Note that speakers F03 and F04 have fewer utterances.}
\label{tab:uaspeech}
\end{table}

\begin{figure}[h!]
    \centering
    \includegraphics[width=1\linewidth]{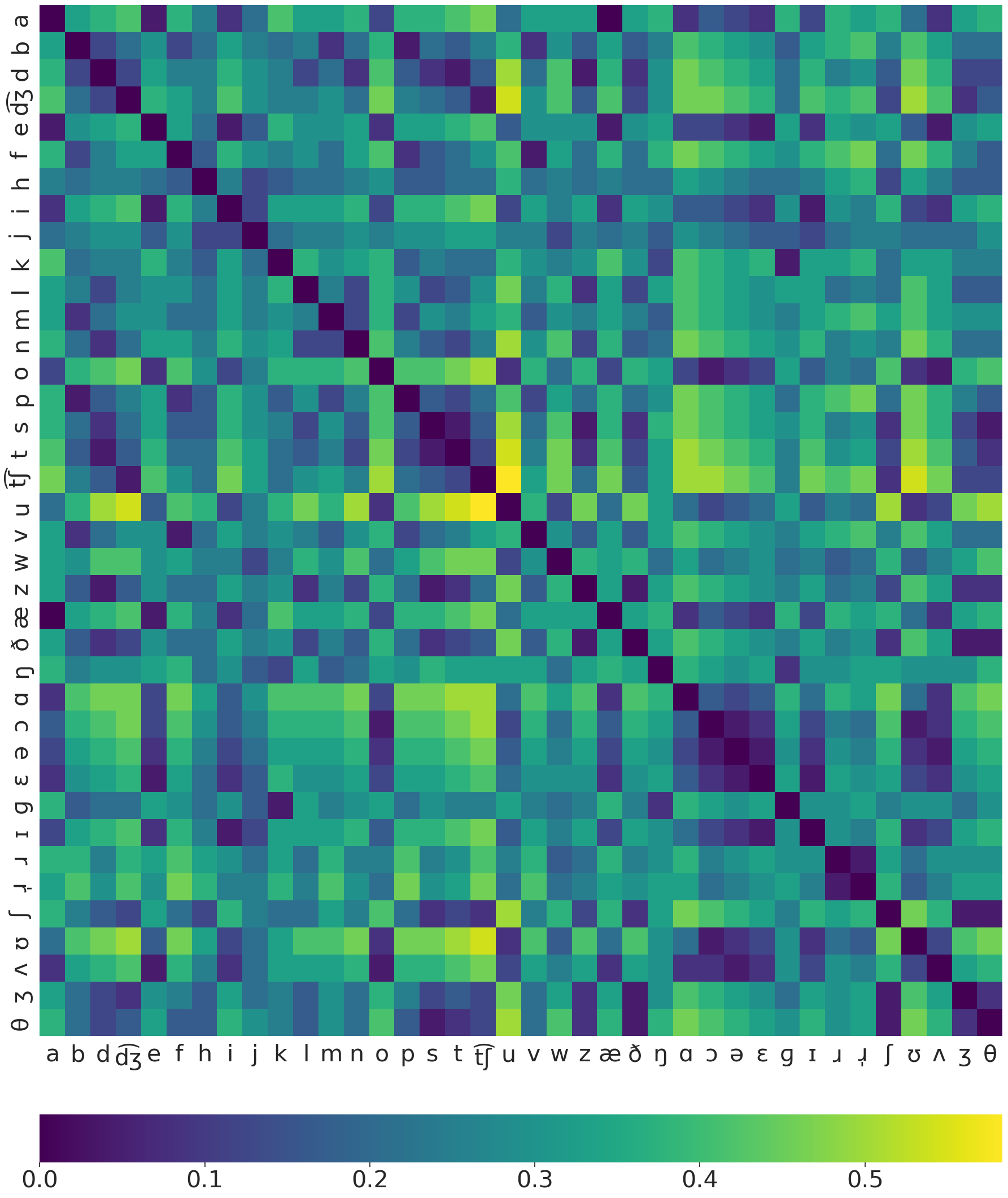}
    \caption{Heat map of phoneme distance matrix (hamming feature edit distance). Brighter areas indicate greater differences in pronunciation between phoneme pair.}
    \label{fig:heatmap}
\end{figure}

\begin{figure}[h!]
    \centering
    \includegraphics[width=1\linewidth]{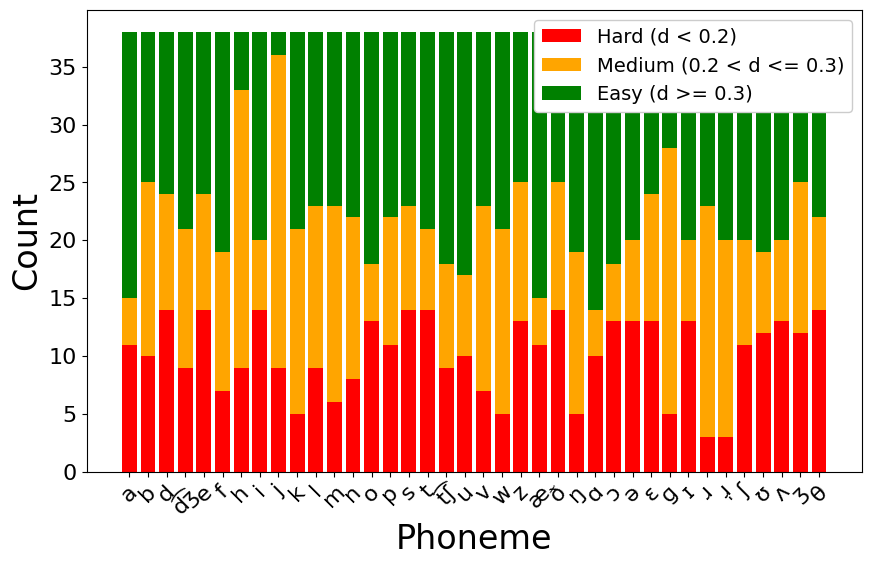}
\caption{Distribution of difficulty of phoneme pairs in 3 levels: Hard ($d < 0.2$), Medium ($0.2 < d \leq 0.3$), and Easy ($d \geq 0.3$). $d$ is phoneme distance between pairs}
    \label{fig:bars}
\end{figure}


\end{document}